\documentclass[pdflatex,sn-mathphys-num]{sn-jnl}% Math and Physical Sciences Numbered Reference Style
\usepackage{graphicx}%
\usepackage{multirow}%
\usepackage{amsmath,amssymb,amsfonts}%
\usepackage{amsthm}%
\usepackage{mathrsfs}%
\usepackage[title]{appendix}%
\usepackage{xcolor}%
\usepackage{textcomp}%
\usepackage{manyfoot}%
\usepackage{booktabs}%
\usepackage{algorithm}%
\usepackage{algorithmicx}%
\usepackage{algpseudocode}%
\usepackage{listings}%
\usepackage{comment}
\theoremstyle{thmstyleone}%
\theoremstyle{thmstyletwo}%

\theoremstyle{thmstylethree}%

\begin{document}

\title[Article Title]{Kinematic Analysis and Integration of Vision Algorithms for a Mobile Manipulator Employed Inside a Self-Driving Laboratory }


\author*[1]{\fnm{Shifa} \sur{Sulaiman}}\email{ssajmech@gmail.com}

\author[1]{\fnm{Tobias Busk} \sur{Jensen}}\email{tobjen21@student.aau.dk}

\author[2]{\fnm{Stefan Hein} \sur{Bengtson}}\email{shbe@create.aau.dk}

\author[1]{\fnm{Simon } \sur{Bøgh}}\email{sibo@es.aau.dk}

\affil*[1]{\orgdiv{Department of Electronic Systems}, \orgname{Aalborg University}, \orgaddress{\street{Fredrik Bajers Vej 7}, \city{Aalborg}, \postcode{9220}, \state{North Denmark Region}, \country{Denmark}}}

\affil[2]{\orgdiv{Department of Architecture and Media Technology}, \orgname{Aalborg University}, \orgaddress{\street{Rendsburggade 14}, \city{City}, \postcode{9000}, \state{North Denmark Region}, \country{Denmark}}}


\abstract{%Recent advances in robotics and autonomous systems have significantly expanded the deployment of robots across a diverse array of applications in lab environments, including automated synthesis, scalable reaction workflows, and collaborative interactions in self-driving laboratories (SDL). 
Recent advances in robotics and autonomous systems have broadened the use of robots in laboratory settings, including automated synthesis, scalable reaction workflows, and collaborative tasks in self-driving laboratories (SDLs).
This paper presents a comprehensive development of a mobile manipulator designed to assist human operators in such autonomous lab environments. Kinematic modeling of the manipulator is carried out based on the Denavit–Hartenberg (DH) convention and inverse kinematics solution is determined to enable precise and adaptive manipulation capabilities. A key focus of this research is enhancing the manipulator’s ability to reliably grasp textured objects as a critical component of autonomous handling tasks. Advanced vision-based algorithms are implemented to perform real-time object detection and pose estimation, guiding the manipulator in dynamic grasping and following tasks. %These perceptual modules work in tandem with the underlying mechanical models to optimize interaction with items of varying shapes, sizes, and material properties without relying on predefined grasp strategies. 
In this work, we integrate a vision method that combines feature-based detection with homography-driven pose estimation, leveraging depth information to represent an object's pose as a 2D planar projection within 3D space. %The robotic system is initially equipped with a structured set of canonical grasp configurations, which define precise spatial relationships between the end effector and objects in standardized poses to ensure stability and reliability during grasping. Once operational, the robot utilizes real-time object recognition and pose estimation to interpret and adjust these grasp configurations. 
This adaptive capability enables the system to accommodate variations in object orientation and supports robust autonomous manipulation across diverse environments.
%The robot is pre-trained with a set of canonical grasps—defined configurations that specify how its end-effector should be positioned relative to an object in a known pose to ensure a secure grasp. Once deployed, the robot can detect objects and estimate their pose in real time, dynamically adapting the pre-trained canonical grasp to match the object's current orientation. Our approach reliably detects well-textured planar objects and estimates their pose with high accuracy, even under moderate out-of-plane rotations. %The integration of visual perception, dynamic control, and model-based manipulation enhances the mobile robot’s dexterity and responsiveness. Whether handling experimental samples, assembling lab components, or conducting automated workflows, the system supports seamless adaptation to complex environments while ensuring safety, stability, and high operational efficiency. 
By enabling autonomous experimentation and human-robot collaboration, this work contributes to the scalability and reproducibility of next-generation chemical laboratories.

%In this study, we introduce a method that combines feature-based detection with homography-driven pose estimation, leveraging depth information to represent an object's pose as a 2D planar projection within 3D space. The robot is pre-trained with a set of canonical grasps—defined configurations that specify how its end-effector should be positioned relative to an object in a known pose to ensure a secure grasp. Once deployed, the robot can detect objects and estimate their pose in real time, dynamically adapting the pre-trained canonical grasp to match the object's current orientation. Our approach reliably detects well-textured planar objects and estimates their pose with high accuracy, even under moderate out-of-plane rotations. 

}

\keywords{Self-Driving Chemical Laboratory, Mobile Manipulator, Kinematic Modeling, Vision-Based Object Detection}



\maketitle

%\section{Introduction}\label{sec1}

\section{Introduction}

Autonomous robotic systems are reshaping research workflows by enabling scalable, efficient, and safe operations in laboratory environments \cite{ref1}. Within this shift, manipulators mounted on mobile platforms  \cite{n1},  \cite{n2}  have emerged as dynamic agents capable of performing tasks like sample transport, equipment handling, and autonomous experimentation. A cornerstone of these capabilities lies in reliable object tracking and grasping, particularly when dealing with fragile or hazardous materials. To perform these tasks with precision, vision-based object detection and pose estimation have become central tools in robotics, allowing for real-time perception and adaptive motion planning. Leveraging color and depth information, robots in self-driving laboratories can interact with diverse objects, adjusting dynamically to changes in shape, texture, and environmental conditions.

While recent innovations have expanded the reach of autonomous systems, deploying robust grasping in unstructured lab environments remains challenging. Factors like lighting variability, occlusion, and clutter introduce complexity to object recognition and manipulation. Overcoming these challenges requires a synergistic framework that combines mechanical modeling, control stability, and perceptual accuracy. In this work, we propose a comprehensive system that integrates kinematic model of a manipulator with real-time perception to enhance grasping reliability in self-driving chemical laboratories. The mobile manipulator is developed to follow and grasp textured objects, supporting autonomous workflows and minimizing human intervention. Major contributions of this paper are as follows:

\begin{itemize}
    \item \textbf{Complete system integration:} Developed a kinematic analysis framework for a mobile manipulator tailored to autonomous chemical laboratories, enabling seamless integration into self-driving lab infrastructure through coordinated control and perception modules.
    
    %\item \textbf{Kinematic analysis:} Use of the Denavit–Hartenberg (DH) convention for systematic kinematic modeling and joint-space behavior prediction.
    
    %\item \textbf{Inverse kinematics (IK) implementation:} Deployment of IK strategies for accurate end-effector positioning during dynamic object tracking and grasping.
    
    %\item \textbf{Dynamic modeling:} Application of Euler–Lagrange formalism to model manipulator dynamics and optimize control under varying payloads.
    
    %\item \textbf{Vision-based perception:} Real-time object detection and pose estimation using RGB-D data to facilitate robust visual servoing for adaptive grasping.
    
    \item \textbf{Simulation and Real-World Performance Assessment :} Validated system performance via simulation and real-world experiments, reporting quantitative metrics such as pose estimation error and grasping success rate to demonstrate feasibility and reliability.
    
    %\item \textbf{Safe handling:} Design of compliant grasping strategies for delicate items such as glass vessels and chemical containers, ensuring operational safety.
\end{itemize}

%A fundamental aspect of this interaction inside a self - driving laboratory is the ability to execute reliable grasping of arbitrary objects, which is essential for effective manipulation and operational efficiency. Consequently, research in robotics has increasingly focused on object detection and pose estimation as key components of dynamic robotic grasping.

Despite substantial progress, object identification and pose recovery remain challenging in real-world laboratory environments due to variability in object shape, texture, and appearance. These challenges are further exacerbated by occlusions, cluttered scenes, inconsistent lighting, and background interference, all of which degrade the reliability of recognition algorithms. To address these issues, we updated a vision pipeline \cite{r1} that integrates feature-based detection with homography-driven pose estimation, augmented by depth sensing. This method enables robust 2D planar projection of objects within 3D space, allowing the mobile manipulator to adapt to dynamic orientations and environmental conditions. By balancing precision with computational efficiency, the system supports real-time manipulation tasks essential for autonomous experimentation and human-robot collaboration in self-driving laboratories.

In this work, we integrate our methodology within a self-driving laboratory setup, where autonomous robotic systems continuously operate without human intervention, leveraging real-time object detection and pose estimation techniques. The self-driving laboratory provides a dynamic environment for validating adaptability and precision in robotic grasping tasks, ensuring that robots can efficiently navigate, recognize, and manipulate objects under varied conditions. By utilizing RGB and depth data within this automated setting, we enhance robotic perception and interaction to follow a textured object and facilitating reliable grasping.

\section{Background}

Object detection has evolved significantly over the past two decades as shown in Table \ref{tab:ref_summary}, transitioning from handcrafted feature-based methods to deep learning-driven architectures. Early approaches relied heavily on sliding window techniques and feature descriptors. One of the foundational methods was the Viola-Jones detector \cite{ref9}, which used Haar-like features and an AdaBoost classifier for real-time face detection. Although efficient, its applicability was limited to rigid objects and constrained environments. The introduction of Histogram of Oriented Gradients (HOG) by Dalal and Triggs \cite{ref10} marked a major advancement in dynamic object detection. HOG features, combined with Support Vector Machines (SVM), provided robust performance in detecting dynamic objects in varied poses and lighting conditions. However, these traditional methods struggled with scale variation and complex backgrounds.

A paradigm shift occurred with the advent of deep learning. Girshick \textit{et al.} \cite{ref11} proposed R-CNN (Regions with Convolutional Neural Networks), which combined region proposals with CNN-based classification. Despite its accuracy, R-CNN was computationally expensive due to its multi-stage pipeline. To address this, Fast R-CNN \cite{ref12} and Faster R-CNN \cite{ref13} introduced architectural refinements, including the Region Proposal Network (RPN), significantly improving speed and efficiency.
Single-stage detectors emerged to further enhance real-time performance. Redmon \textit{et al.} \cite{ref14} introduced YOLO (You Only Look Once) in 2016, which reframed object detection as a regression problem. YOLO’s unified architecture enabled real-time detection but initially suffered from localization errors. Successive versions like YOLOv2, YOLOv3, and YOLOv4 improved accuracy and robustness, while YOLOv5 and YOLOv7 incorporated advanced backbone networks and data augmentation strategies.
Simultaneously, Liu \textit{et al.} \cite{ref15} proposed SSD (Single Shot MultiBox Detector), which used multi-scale feature maps for detecting objects of varying sizes. RetinaNet \cite{ref16} tackled the class imbalance problem using Focal Loss, achieving high accuracy on challenging datasets like COCO.

Transformer-based models have recently gained traction. Carion \textit{et al.} \cite{ref17} introduced a novel end-to-end detection framework named 'DETR' using transformers for object localization and classification. Although the method simplified the pipeline, its convergence time was a concern. Subsequent models like Deformable DETR and DINO refined the architecture for faster training and improved performance. Despite these advancements, several challenges persist. Small object detection, occlusion handling, and domain adaptation remain active research areas. Context-aware models \cite{ref18} and zero-shot detection frameworks are being explored to address these limitations. Moreover, real-time deployment on edge devices demands lightweight models like YOLOv10 \cite{ref19} which optimize both accuracy and latency. 

SIFT (Scale-Invariant Feature Transform) \cite{refSIFT1} is a robust computer vision technique used for detecting and describing local features in images. It enables reliable object recognition by extracting keypoints that are invariant to scale, rotation, and partial affine transformations, making it highly effective in dynamic or cluttered environments. These features are matched across images using descriptor-based indexing methods, allowing for precise localization and identification of objects.  Vaithiyanathan \textit{et al.} \cite{refSIFT} presented a real-time object recognition system that leveraged SIFT-based local image features, which are resilient to rotation, scaling, translation, and partial illumination changes. Keypoints are extracted using phased filtering in scale space and matched via nearest-neighbor search to determine object presence. Recognition results, based on descriptor count thresholds, are displayed on a microcontroller-driven output unit.

Object pose estimation \cite{pose3} refers to the process of determining the position and orientation of an object in a given coordinate frame, typically in 3D space. It plays a critical role in robotics, augmented reality, and computer vision applications where spatial understanding is essential. Pose estimation algorithms often rely on visual cues such as keypoints, contours, or depth data to infer the object's six degrees of freedom (translation along x, y, z axes and rotation around them). Techniques range from classical methods like Perspective-n-Point (PnP) and template matching to deep learning-based approaches that regress pose directly from image features. Accurate pose estimation enables tasks like grasping, navigation, and interaction with objects in dynamic environments. Object pose estimation is a cornerstone of intelligent robotic perception, enabling machines to understand and interact with their environment in a spatially aware manner. By accurately determining an object's position and orientation, robots can perform tasks such as grasping, manipulation, navigation, and assembly with precision and reliability \cite{pose4}. This capability is especially critical in dynamic or unstructured settings like warehouses, homes, or surgical environments where real-time decision-making depends on spatial context. Pose estimation also facilitates seamless integration between vision and control systems, allowing robots to adapt to changing conditions, avoid collisions, and execute complex motions. Ultimately, it transforms raw sensory data into actionable intelligence, bridging perception and action in autonomous systems.

Wang \textit{et al.} \cite{pose1} presented DenseFusion, a real-time framework for 6D object pose estimation from RGB-D images. The method introduced a heterogeneous architecture that processed RGB and depth data separately, then fused them using a dense pixel-wise feature embedding network to estimate object poses. By integrating an end-to-end iterative refinement module, DenseFusion significantly improved pose accuracy in cluttered scenes and demonstrates strong performance on benchmark datasets and real-world robotic manipulation tasks. Xiang \textit{et al.} \cite{pose2} proposed PoseCNN, a convolutional neural network designed for 6D object pose estimation using RGB images. The method estimated 3D translation by localizing object centers and predicting depth, while 3D rotation is regressed via quaternion representation, with a novel loss function to handle symmetric objects. PoseCNN demonstrated strong robustness to occlusion and achieves state-of-the-art results on the OccludedLINEMOD dataset, supported by the introduction of the large-scale YCB-Video dataset for benchmarking.

The rapid advancement of object detection and pose estimation has not only revolutionized visual perception systems but also significantly empowered robotic platforms, particularly mobile manipulators. These systems rely on robust and real-time object detection to perceive, localize, and interact with their surroundings in dynamic and unstructured environments. Mobile manipulators that integrate a versatile robotic arm with a mobile base have become indispensable in a range of domains such as industrial automation, logistics, and healthcare. These integrated platforms combine mobility with dexterous manipulation, enabling them to perform complex tasks in dynamic environments. Their emerging role in chemistry laboratories presents unique challenges and opportunities, especially given the sensitive and often hazardous nature of chemical substances. Tasks such as transferring fragile glassware, dispensing reagents, or interacting with laboratory equipment demand exceptional precision and reliability.

%Angelopoulos, Anastasios, Matthew Verber, Collin McKinney, James Cahoon, and Ron Alterovitz. “High-Accuracy Injection Using a Mobile Manipulation Robot for Chemistry Lab Automation.” IEEE/RSJ International Conference on Intelligent Robots and Systems (IROS), 2023. https://robotics.cs.unc.edu/publications/Angelopoulos2023_IROS.pdf

%Fizet, Julien. “Lab Automation with Mobile Robots.” Antdriven Insights, 2025. https://insights.antdriven.com/lab-automation-mobile-robots

%Robotnik Automation. “Uses of a Mobile Manipulator Robot: Beyond Robotic Arms.” Robotnik Blog, 2025. https://robotnik.eu/uses-of-a-mobile-manipulator-robot-beyond-robotic-arms/

%CRL Solutions. “Telemanipulators for Radiopharmaceutical Laboratories.” CRL Technical Overview, 2024. https://crlsolutions.com/products/telemanipulators/

%PS Lift. “Vacuum Manipulators in the Chemical and Pharmaceutical Industry.” PS Lift Product Guide, 2024. https://en.ps-lift.com/product/vacuum-manipulator-in-chemical-and-pharmaceutical-industry/

%Dalmec North America. “How Industrial Manipulator Solutions Benefit the Chemical Industry.” Dalmec NA Blog, 2024. https://www.dalmec-na.com/blog/how-industrial-manipulator-solutions-benefit-the-chemical-industry/

%Debijadi, M. L. (2023). Simulation design of a robotic mobile manipulator for Material Acceleration Platforms (Master’s thesis, Aalborg University). Aalborg University Student Projects.

Despite the growing interest in automating laboratory workflows, deploying mobile manipulators and fixed base manipulators \cite{c1}, \cite{c2} in chemical and pharmaceutical environments presents several persistent challenges including precision handling, infrastructure variability, and safety in hazardous conditions. Recent studies have begun addressing these gaps through targeted innovations. Angelopoulos et al. \cite{ref2} tackled the challenge of high-precision sample injections by developing a mobile manipulation framework that combines deep learning-based syringe localization with visual servoing, achieving millimeter-scale accuracy in gas chromatograph interactions. While robust to navigation and grasping uncertainties, the study acknowledged limitations in generalizing across diverse lab setups, pointing to future work in multi-modal sensing and broader task automation.

Fizet et al. \cite{ref3} and JAG Robotics \cite{ref55} addressed logistical inefficiencies by deploying mobile manipulators equipped with vision systems and temperature-controlled storage for inter- and intra-laboratory transport. Their work improved traceability and operational throughput, yet highlighted the difficulty of integrating robots across heterogeneous lab infrastructures underscoring the need for standardized interfaces and coordinated fleet management. In hazardous environments, the importance of task-specific end-of-arm tooling (EOAT) was emphasized by \cite{ref4}, which identified a gap in adaptive control systems capable of handling varied chemical containers and environmental constraints. CRL Solutions \cite{ref5} contributed to safety in radiopharmaceutical production through telemanipulators for remote vial handling, but noted limitations in automation scalability and integration with modern robotic platforms.

Industrial manipulators such as those from PS Lift \cite{ref6} and Dalmec NA \cite{ref7} have improved ergonomic material handling in chemical manufacturing. However, their static and task-specific nature limits adaptability, suggesting future research into modular designs and sensor-integrated systems for intelligent automation. Debijadi \cite{ref8} explored simulated deployment using ROS2 and MoveIt2 for autonomous navigation and manipulation within lab settings. While promising, the study emphasized the need for real-world validation and robust control under dynamic conditions. Collectively, these works demonstrate meaningful progress in addressing key deployment challenges of mobile manipulators in chemical labs. Yet, gaps remain in generalization, adaptive control, and system integration pointing to future opportunities in multi-modal perception, standardized protocols, and hybrid autonomy frameworks.

\begin{table}[h]
    \centering
    \caption{Comparison of object detection methods}
    \label{tab:ref_summary}
    
    \begin{tabular}{|p{2cm}|p{3cm}|p{3cm}|p{3cm}|}
        \hline
        \textbf{Method} & \textbf{Key Features} & \textbf{Strengths} & \textbf{Limitations} \\ \hline
        Viola--Jones \cite{ref9} & Haar-like features, AdaBoost classifier & Real-time for rigid objects (e.g., faces) & Limited to constrained settings, poor generalization \\ \hline
        HOG + SVM \cite{ref10} & Gradient-based descriptors & Robust to human detection, lighting variations & Struggles with scale variation and cluttered scenes \\ \hline
        R-CNN \cite{ref11} & Region proposals + CNN classification & High accuracy & Slow, multi-stage pipeline \\ \hline
        Fast/Faster R-CNN \cite{ref12,ref13} & Shared CNN backbone, Region Proposal Network & Faster, good accuracy & Still computationally heavy \\ \hline
        YOLO (v1--v7) \cite{ref14} & Single-stage regression & Real-time detection, simple pipeline & Early versions had localization errors \\ \hline
        SSD \cite{ref15} & Multi-scale feature maps & Good for varied object sizes & Lower accuracy than two-stage models \\ \hline
        RetinaNet \cite{ref16} & Focal loss & Strong accuracy, handles class imbalance & Higher latency than YOLO \\ \hline
        DETR / Transformer-based \cite{ref17} & End-to-end transformer detection & Simplifies pipeline, strong accuracy & Long training time, resource-intensive \\ \hline
        YOLOv10 \cite{ref19} & Lightweight, optimized for edge devices & Balance of speed and accuracy & Still evolving, limited benchmarks \\ \hline
    \end{tabular}
\end{table}

\section{Methodology}
We have carried out kinematic modeling and performed workspace analysis to determine the feasible interaction zone. 
Inverse kinematic solutions obtained using kinematic equations are applied to compute joint configurations that allowed the system to follow the object’s trajectory to reach object pose with precision, facilitating reliable grasping and manipulation.
We employed a planar pose estimation algorithm for tracking a textured object and later the determined pose information is used for planning trajectory. Initially, the process begins with feature extraction and template matching, ensuring robust identification of key object characteristics. This is followed by homography estimation and perspective transformation, which allow for accurate spatial mapping and alignment. Subsequently, directional vectors are estimated on the object surface to facilitate precise orientation analysis. Finally, planar pose estimation is performed utilizing depth data, enabling reliable grasping and interaction.

The object detection approach started by extracting distinctive features using SIFT from images of planar objects and matching them with features detected in images captured by the camera. These features, such as edges, corners, blobs, and ridges, serve as key patterns for describing the image content. Following feature extraction, the presence of an object is ascertained by evaluating the correspondence between extracted features and those of a reference image associated with the target object. %For non-binary descriptors like SIFT and SURF, we use the Nearest Neighbor algorithm \cite{1}, though its high computational cost can hinder real-time pose updates with multiple objects. 
To mitigate the issue with high computational cost, we employed the FLANN implementation \cite{3} of K-d Nearest Neighbor Search, optimizing matching for high-dimensional features. 
%For binary descriptors such as AKAZE and BRISK, we apply the Hamming distance ratio method. If at least ten matches are found, the object is considered present in the scene.
To estimate the homography, we used matches obtained from the nearest neighbor search as input. However, some of these matches can be incorrect, leading to false correspondences that hinder accurate homography estimation. To address this, we employed RANSAC \cite{4}, which robustly estimated homography by selecting only inlier matches. %Unlike conventional methods that incorporate large datasets before eliminating outliers, RANSAC showcases a more efficient approach by using a minimal subset of observations to iteratively refine model parameters and detect outliers through random sampling. 
Unlike conventional approaches that process large datasets before attempting to eliminate outliers often require substantial computational resources and risking model distortion, RANSAC adopts a more efficient strategy. It begins with a minimal, randomly selected subset of observations and iteratively refines the model parameters by evaluating the consistency of additional data points. Through this randomized sampling and consensus-building mechanism, RANSAC effectively isolates outliers early in the process, enabling accurate homography estimation even in the presence of noisy or ambiguous data.
This optimization enhanced both speed and accuracy, making it well-suited for real-time applications. In this study, perspective transformation is utilized to approximate corresponding points within the test image template, facilitating the derivation of basis vectors for the object's surface. Subsequently, depth information is incorporated to compute the surface normal of the planar object, thereby enabling accurate estimation of its 3D pose.

\begin{figure}[hbt!]
    \centering
    \includegraphics[width=1\textwidth]{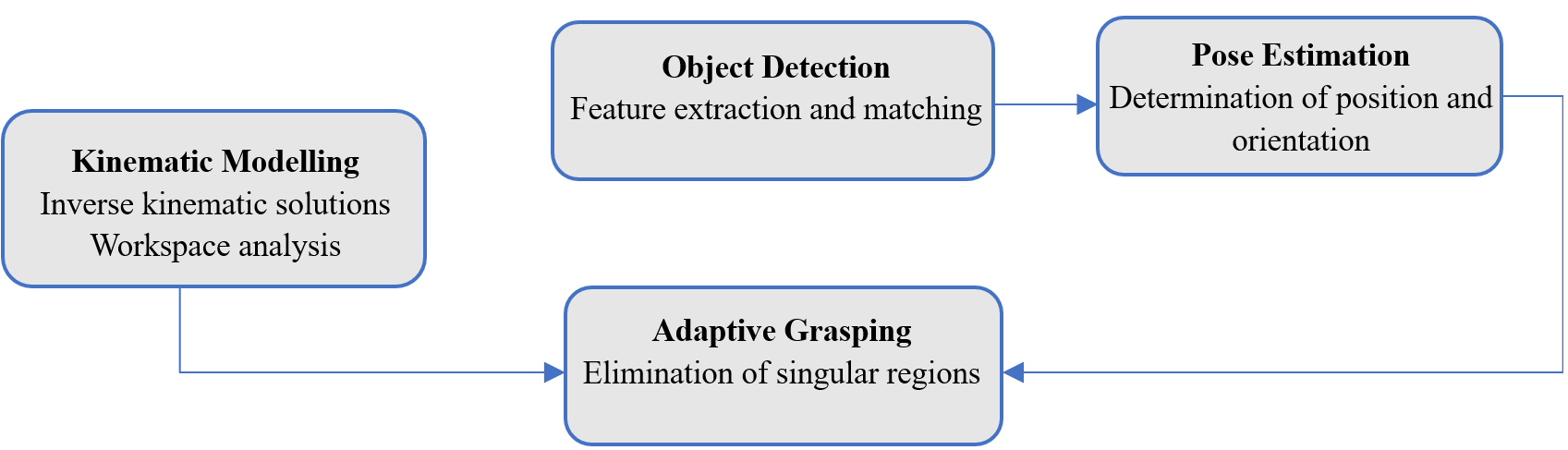}
    \caption{Methodology adopted in this work}
    \label{flowchart}
\end{figure}
The flow of different tasks carried out in this work is shown in Fig.~\ref{flowchart}. The process of robotic manipulation begins with kinematic modeling, which involved solving inverse kinematics and conducting workspace analysis to determine feasible configurations for the robot’s end effector. This foundational stage ensured that the manipulator can reach and interact with objects within its operational domain. Following this, object detection is performed through feature extraction and matching techniques, enabling the system to identify and localize target objects within the environment. Once detected, pose estimation is employed to determine the precise position and orientation of the object, providing critical spatial information for manipulation. The final stage involved adaptive grasping, in which the robot dynamically adjusts its grasp strategy to eliminate singular regions and ensure stable interaction with the object. This sequential framework facilitates reliable and flexible autonomous handling in complex and variable settings.

\subsection{Description of the mobile manipulator}
\begin{figure}[hbt!]
    \centering
    \includegraphics[width=0.8\textwidth]{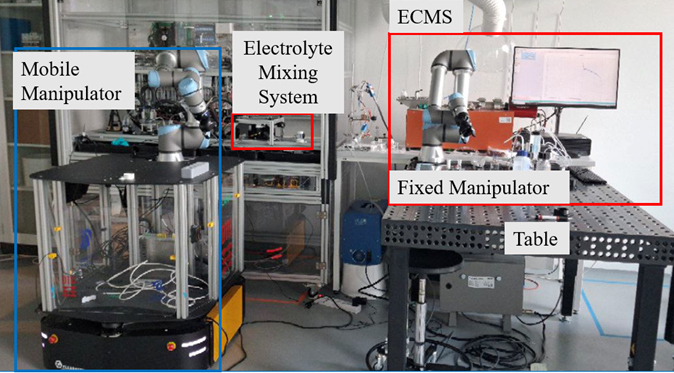}
    \caption{Laboratory setup}
    \label{lab}
\end{figure}

\begin{figure}[hbt!]
    \centering
\includegraphics[width=0.6\textwidth, height = 2.2 in]{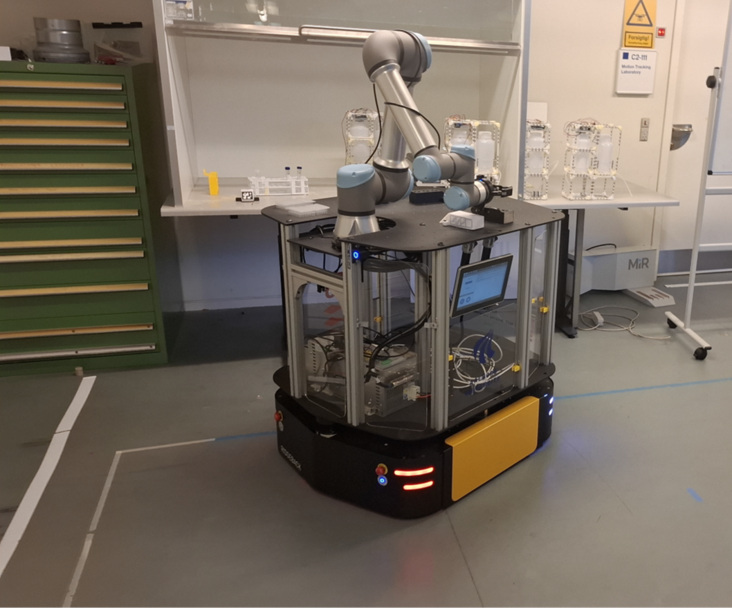} 
    \caption{Mobile manipulator}
   \label{fig.1}
\end{figure}

This study adopted a structured methodology to design, implement, and evaluate a mobile manipulator robot integrated within a self-driving chemistry laboratory. The development process begins with the design of a mobile robotic system capable of autonomous navigation and execution of pick-and-place operations in a dynamic laboratory setting, as illustrated in Fig.~\ref{lab}. 
The experimental environment comprised of a mobile manipulator, a fixed-base manipulator, an electrolyte mixing system, and an Electrochemical Mass Spectrometry (ECMS) unit. The mobile manipulator system is built upon the Ridgeback omni-directional platform from Clearpath Robotics, outfitted with a Universal Robots UR5e collaborative robotic arm and a Robotiq Hand-E adaptive gripper as shown in  Fig.~\ref{fig.1}. The Ridgeback base offers holonomic motion and high payload capacity, enabling precise and agile maneuvering within cluttered laboratory spaces. Mounted on the platform, the UR5e arm provides 6 degrees of freedom (DOF), high positional accuracy, and integrated force-torque sensing, making it suitable for delicate manipulation tasks. 

At the end-effector, the Robotiq Hand-E gripper delivers versatile grasping capabilities, accommodating a broad range of laboratory objects from small vials to larger containers. This integrated configuration enabled the mobile manipulator to autonomously traverse between workstations, perform object handling, manage chemical samples, and execute tasks such as equipment calibration and maintenance. To support intelligent operation, the system is equipped with two LiDAR sensors and three Intel RealSense cameras, providing rich spatial and visual data. Software frameworks including ROS (Robot Operating System) and MoveIt are employed to facilitate real-time localization, motion planning, and obstacle avoidance. Together, these components form a robust and adaptable platform for autonomous experimentation and laboratory automation.

\subsection{Kinematic Modeling of the Mobile Manipulator}
Kinematic modeling of the mobile manipulator involved the systematic representation of its motion capabilities through mathematical formulations, independent of dynamic forces. This process integrated the kinematic chains of both the mobile base and the manipulator arm to determine the position and orientation of the end-effector relative to a global reference frame. The model encompassed forward and inverse kinematics, as well as the derivation of the Jacobian matrix to relate joint velocities to end-effector motion. Such modeling is fundamental for enabling precise motion planning, control algorithms, and coordinated task execution in complex environments.
\subsubsection{Kinematic modeling of the mobile platform}
\begin{figure}[hbt!]
    \centering
\includegraphics[width=0.4\textwidth, height = 2 in]{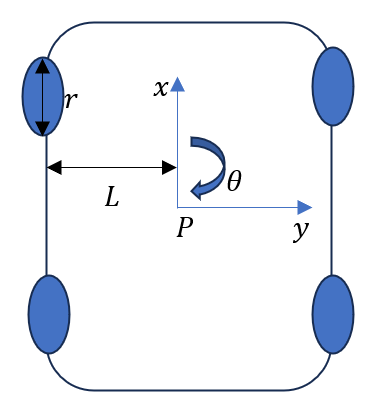} 
    \caption{Representation of mobile base}
   \label{fig.rep}
\end{figure}

An omni-directional mobile robot shown in Fig.\ref{fig.rep} equipped with 4 omni wheels provides enhanced maneuverability compared to conventional wheeled robots, allowing for instantaneous translation and rotation in any direction. To accurately model its kinematics, we define equations governing its motion based on robotic mechanics principles, considering wheel velocities, forces, and torques. The kinematic model describes the relationship between wheel velocities and the platform’s linear/angular velocity. %Each omni-wheel allows motion in two directions: one from direct rotation of the wheel and another due to the passive rolling effect perpendicular to the wheel rotation axis. The mobile robot platform features a holonomic base with four omnidirectional wheels mounted at the corners of a rectangular chassis. 
Each wheel has a radius of $r$ and is positioned at a distance $L$ from the robot’s geometric center, labeled as point $P$. This symmetric configuration allows the robot to achieve full planar mobility, enabling independent control of translation and rotation.
The robot pose, P is defined as given in Eq. \ref{p}
\begin {equation}
P = \begin{bmatrix} x  ~y  ~\theta \end{bmatrix}^T
\label{p}
\end{equation}
where \( x, y \) represent the global position, and \( \theta \) is the yaw angle. The velocity vector is given in Eq. \ref{q}
\begin{equation}
V = \begin{bmatrix} v_x  ~v_y  ~\omega \end{bmatrix}^T
\label{q}
\end{equation}
where \( v_x, v_y \) are the linear velocities, and \( \omega \) is the angular velocity. The relationship between the wheel velocities \( v_1, v_2, v_3, v_4 \), and the platform velocity is given in Eq. \ref{r}
\begin {equation}
\begin{bmatrix} v_1  ~v_2  ~v_3 ~v_4 \end{bmatrix}^T = J^{-1} \begin{bmatrix} v_x ~v_y ~\omega \end{bmatrix}^T
\label{r}
\end{equation}
where \( J^{-1} \) is the inverse Jacobian matrix as given in Eq. \ref{n}
\begin {equation}
J^{-1} =
\begin{bmatrix}
\frac{1}{r} & -\frac{1}{r} & -\frac{L}{r} \\
\frac{1}{r} & \frac{1}{r} & L/r \\
\frac{1}{r} & -\frac{1}{r} & L/r \\
\frac{1}{r} & \frac{1}{r} & -L/r
\end{bmatrix}
\label{n}
\end{equation}
%where \( r \) is the wheel radius, and \( L \) is the distance from the robot center to each wheel.
Using this formulation, wheel velocities can be computed from the desired robot velocity, enabling effective motion control of the omni-directional platform.

%\section{Kinematic Modeling of the UR5e}
\subsubsection{Kinematic Modeling of the UR5e}

The UR5e, developed by Universal Robots, is a 6-DoF articulated robotic manipulator designed for versatile industrial and research applications. Notable for its compact structure, lightweight design, and user-friendly programming interface, the UR5e also incorporates advanced safety features that facilitate collaborative operation in dynamic environments. A distinctive aspect of its architecture is the non-coincidental arrangement of the final 3 joints, which do not form a traditional spherical wrist. Consequently, all six joints contribute independently to both translational and rotational motion of the end-effector, thereby increasing the manipulator’s dexterity. However, this configuration introduces additional complexity in kinematic modeling, as the decoupling of position and orientation typically exploited in manipulators with coincidental wrists is not directly applicable. The kinematic analysis of the UR5e requires a comprehensive formulation of both forward and inverse kinematics, often employing Denavit–Hartenberg parameters to systematically represent joint transformations. This modeling is essential for accurate trajectory planning, control, and integration into autonomous robotic systems.

%\subsection{Kinematic Parameters of UR5}

\begin{figure}[hbt!]
    \centering
\includegraphics[width=0.4\textwidth, height = 4 in]{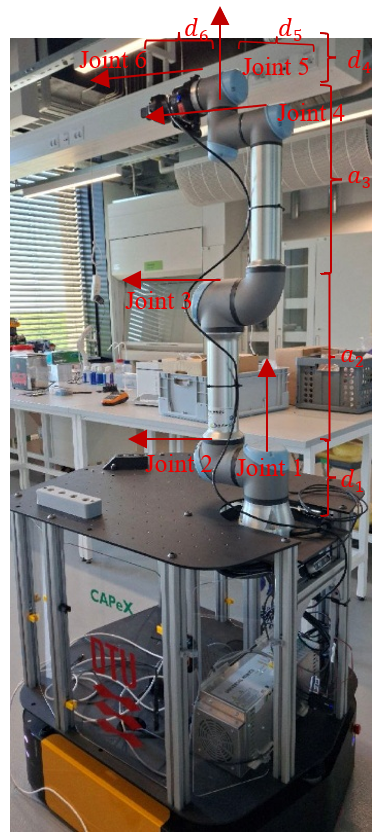} 
    \caption{Mobile manipulator}
   \label{fig.manipulator}
\end{figure}

Figure~\ref{fig.manipulator} depicts the schematic of the UR5 and the coordinate frame assignment for each joint, based on which the Denavit-Hartenberg (DH) parameters are defined. The DH parameters are listed in Table~\ref{tab:dh_params}.

\begin{table}[h]
\centering
\caption{DH parameters of UR5}
\label{tab:dh_params}
\begin{tabular}{@{}c|cccc@{}}
\toprule
$i$ & $a_i$ (m) & $d_i$ (m) & $\alpha_i$ (rad) & $\theta_i$ \\ \midrule
1 & 0 & $d_1 = 0.08916$ & $\pi/2$   & $\theta_1$ \\
2 & $a_2 = -0.425$ & 0 & 0 & $\theta_2$ \\
3 & $a_3 = -0.39225$ & 0 & 0 & $\theta_3$ \\
4 & 0 & $d_4 = 0.10915$ & $\pi/2$ & $\theta_4$ \\
5 & 0 & $d_5 = 0.09456$ & $-\pi/2$ & $\theta_5$ \\
6 & 0 & $d_6 = 0.0823$ & 0 & $\theta_6$ \\ \bottomrule
\end{tabular}
\end{table}

%\subsection{Forward Kinematics}
The transformation matrix, $T$ is obtained as given in Eq. \ref{t}
\begin{equation}
    T = 
    \begin{bmatrix}
        R_e & P_e \\ 0 & 1
    \end{bmatrix}
   \label{t} 
\end{equation}
where rotation matrix, $R_e$ is obtained as given in Eq. \ref{rotation}

\begin{equation}
R = 
\begin{bmatrix}
n_x & o_x & a_x \\
n_y & o_y & a_y \\
n_z & o_z & a_z
\end{bmatrix}
\label{rotation}
\end{equation}
Using transformation matrices based on DH parameters, the end-effector position $P_e$ $(P_x, P_y, P_z)$ is computed as follows:
\begin{align}
P_x &= d_5 \cos(\theta_1)\sin(\theta_{234}) + d_4 \cos(\theta_1) - d_6 \cos(\theta_1)\cos(\theta_{234})+ a_2 \cos(\theta_1)\cos(\theta_2) +\nonumber \\
&\quad d_6 \cos(\theta_5)\sin(\theta_1) + a_3 \cos(\theta_1)\cos(\theta_2)\cos(\theta_3) - a_3 \cos(\theta_1)\sin(\theta_2)\sin(\theta_3) \\
P_y &= d_5 \sin(\theta_1)\sin(\theta_{234}) - d_4 \sin(\theta_1) - d_6 \sin(\theta_1)\cos(\theta_{234}) - d_6 \cos(\theta_1)\cos(\theta_5) + \nonumber \\
&\quad a_2 \sin(\theta_1)\cos(\theta_2)  + a_3 \sin(\theta_1)\cos(\theta_2)\cos(\theta_3) - a_3 \sin(\theta_1)\sin(\theta_2)\sin(\theta_3) \\
P_z &= d_1 - d_6 \sin(\theta_{234})\cos(\theta_5) + a_3 \sin(\theta_2)\cos(\theta_3)  + a_3 \cos(\theta_2)\sin(\theta_3) + a_2 \sin(\theta_2) \nonumber \\
&\quad - d_5 \cos(\theta_{234})
\end{align}
where, $\theta_{234} = \theta_2 + \theta_3 + \theta_4$.
%\subsection{Inverse Kinematics}
To solve inverse kinematics, we find a joint configuration vector $\mathbf{q} = [\theta_1, \theta_2, ..., \theta_6]^T$ such that the robot achieves the desired pose. The solution begins by isolating $\theta_1$ using the position coordinates of joint 5 as given in Eq. \ref{eq.1}.
\begin{equation}
-\sin(\theta_1)(P_x - d_6 n_x) + \cos(\theta_1)(P_y - d_6 n_y) = -d_4
\label{eq.1}
\end{equation}
Upon solving the Eq. \ref{eq.1} , we obtain:

\begin{equation}
\theta_1 = \text{atan2}(\text{ps}_y, \text{ps}_x) \pm \cos^{-1}\left(\frac{d_4}{\sqrt{\text{ps}_x^2 + \text{ps}_y^2}}\right)
\end{equation}
Similar formulations follow to compute $\theta_5$ and $\theta_6$ by comparing direction vectors of relevant transformation axes. The remaining joint angles $\theta_2$, $\theta_3$, and $\theta_4$ are determined as a planar 3-RRR mechanism solution.
%\subsubsection{Solving for \texorpdfstring{$\theta_2$, $\theta_3$, $\theta_4$}{θ₂, θ₃, θ₄}}
After computing $\theta_1$, we transformed the desired end-effector position into the coordinate frame of joint 2. Let $r_s$ denote the distance from joint 2 to the wrist center. This value can be determined by applying geometric relationships as follows:

\begin{align*}
r_s &= \sqrt{P_x^2 + P_y^2 + (P_z - d_1)^2} \\
\phi &= \tan^{-1}\left(\frac{P_z - d_1}{\sqrt{P_x^2 + P_y^2}}\right)
\end{align*}

Using the law of cosines, we can determine $\theta_3$ as follows:

\begin{align*}
\cos(\theta_3) &= \frac{r_s^2 - a_2^2 - a_3^2}{2a_2 a_3} \\
\theta_3 &= \cos^{-1}\left(\frac{r_s^2 - a_2^2 - a_3^2}{2a_2 a_3}\right)
\end{align*}

We can derive $\theta_2$ based on $\theta_3$ as follows:

\begin{align*}
\theta_2 &= \phi - \tan^{-1}\left(\frac{a_3 \sin(\theta_3)}{a_2 + a_3 \cos(\theta_3)}\right)
\end{align*}

\begin{figure}[hbt!]
    \centering
    \includegraphics[width=1\textwidth]{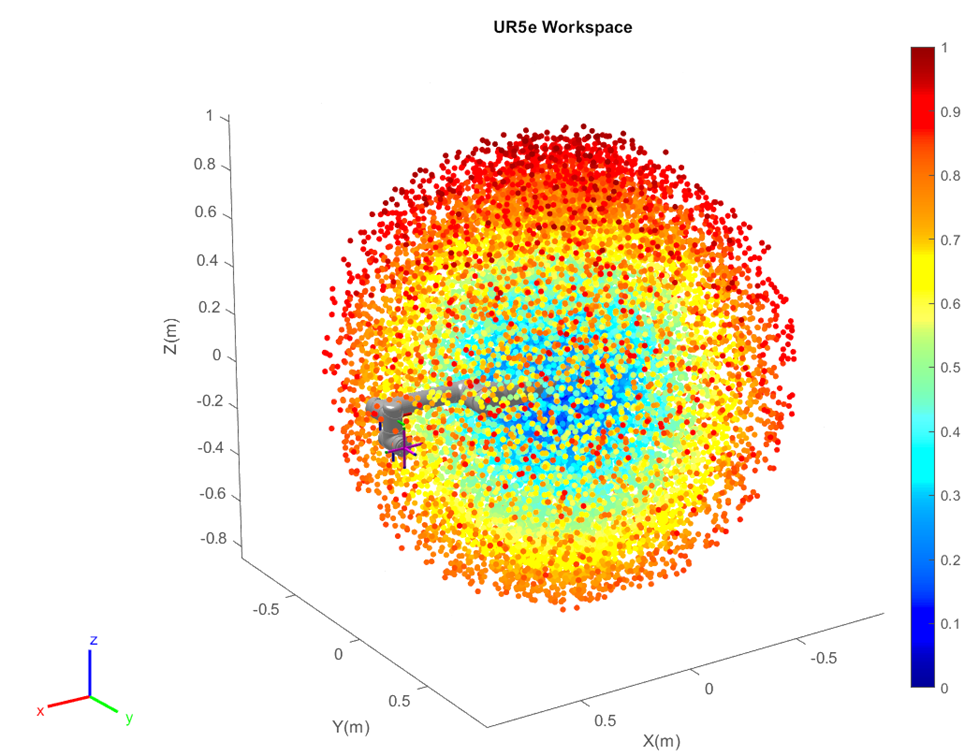}
    \caption{Reachable workspace of the UR5e manipulator. The 3D plot illustrates the distribution of end-effector positions, with color indicating normalized reachability values, highlighting the arm’s effective operational volume.}
    \label{workspace}
\end{figure}

Finally, the value of $\theta_4$ is obtained based on Eq. \ref{eq.51}

\begin{align}
\theta_4 &= \theta_{234} - \theta_2 - \theta_3
\label{eq.51}
\end{align}
%Here, $\theta_{234}$ is computed from the orientation part of the transformation matrix.
%\subsubsection{Solving for \texorpdfstring{$\theta_5$}{θ₅} and \texorpdfstring{$\theta_6$}{θ₆}}
From the rotational part of the transformation matrix, we can compute $\theta_5$ as follows:
\begin{equation}
\theta_5 = \cos^{-1}\left(\frac{P_x \sin(\theta_1) - P_y \cos(\theta_1) - d_4}{d_6}\right)
\end{equation}
The wrist orientation yields $\theta_6$ as follows:
\begin{equation}
\theta_6 = \tan^{-1}\left(\frac{n_y \cos(\theta_1) - n_x \sin(\theta_1)}{o_y \cos(\theta_1) - o_x \sin(\theta_1)}\right)
\end{equation}
%where $n$ and $o$ vectors correspond to the orientation axes extracted from the end-effector pose.
The workspace of the UR5e manipulator was plotted based on the Eqs. 7 - 9 as shown in Fig.\ref{workspace}. The manipulator’s workspace spans approximately $2.27~\text{m}^3$, defined by its $850~\text{mm}$ reach and constrained by joint limits and end-effector configuration.

\section{Vision-Based Grasping Framework}

The vision algorithm \cite{r1} enabled robust object manipulation by integrating visual perception with adaptive grasp planning. The framework consisted of two primary components as follows:

\begin{itemize}
  \item Object detection and planar pose estimation
  \item Adaptive grasping plan estimation
\end{itemize}

\subsection{Object Detection and Pose Estimation}

Accurate pose estimation is essential for reliable robotic manipulation. The proposed method comprised four sequential stages:

\begin{enumerate}
  \item Feature extraction and descriptor matching
  \item Homography computation and perspective transformation
  \item Estimation of object coordinate frame via directional vectors
  \item Pose refinement using depth information
\end{enumerate}
%\paragraph{Feature Extraction and Descriptor Matching}
Planar objects were identified using the Scale-Invariant Feature Transform (SIFT) feature detection algorithm, which extracted unique keypoints and descriptors from images to facilitate accurate recognition and matching. These algorithms extracted keypoints and generated descriptors that encode local image structure. Matching was performed using FLANN for floating-point descriptors or Hamming distance for binary descriptors, establishing correspondences between the input image and a reference template.
%\paragraph{Homography Estimation and Perspective Transformation}
Using matched keypoints, a homography matrix $\mathbf{H}$ was computed to model the planar transformation as follows:
\begin{equation}
\begin{bmatrix}
a \\ 
b \\ 
c
\end{bmatrix}
=
\mathbf{H}
\begin{bmatrix}
x \\ 
y \\ 
1
\end{bmatrix}
\end{equation}
The transformed coordinates $(x', y')$ are derived as:
\begin{equation}
x' = \frac{a}{c}, \quad y' = \frac{b}{c}
\end{equation}
RANSAC was applied to filter outliers and ensure robust homography estimation.
%\paragraph{Directional Vector Estimation}
To define the object’s local coordinate frame, three reference points were selected as follows:
\begin{equation}
P_c = (w/2, h/2), \quad P_x = (w, h/2), \quad P_y = (w/2, 0)
\end{equation}
where, $w$ and $h$ represent width and height of the object respectively.
These points were projected into 3D using the aligned RGB-D data from the RealSense camera, yielding vectors:
\begin{equation}
\vec{i} = \frac{\vec{x}}{\|\vec{x}\|}, \quad \vec{j} = \frac{\vec{y}}{\|\vec{y}\|}, \quad \vec{k} = \frac{\vec{x} \times \vec{y}}{\|\vec{x} \times \vec{y}\|}
\end{equation}
The orthonormal basis $(\vec{i}, \vec{j}, \vec{k})$ defined the object’s orientation in space.
%\paragraph{Pose Estimation Using Euler Angles}
The rotation matrix $R$ was constructed from the directional vectors:
\begin{equation}
R =
\begin{bmatrix}
i_x & j_x & k_x \\
i_y & j_y & k_y \\
i_z & j_z & k_z
\end{bmatrix}
\end{equation}
Euler angles $(\phi, \theta, \psi)$ were computed as follows:
\begin{equation}
\theta = \tan^{-1}(j_z), \quad \phi = \sin^{-1}(-i_z), \quad \psi = \tan^{-1}\left(\frac{i_y}{i_x}\right)
\end{equation}
These angles represented the object’s orientation with respect to the reference frame and were used for grasp planning.
%\subsection{Adaptive Grasp Generation}
Grasping was achieved by computing a transformation between the object pose and the gripper pose. The grasp transformation $\mathbf{T}_g$ was recorded as follows:
\begin{equation}
\mathbf{T}_g = \mathbf{T}_o^{-1} \cdot \mathbf{T}_b
\end{equation}
Here, $\mathbf{T}_o$ is the object pose and $\mathbf{T}_b$ is the gripper pose. During inference, the grasp pose was adapted using the updated object pose $\mathbf{T}_o'$ as follows:
\begin{equation}
\mathbf{T}_g' = \mathbf{T}_o' \cdot \mathbf{T}_g
\end{equation}
The final grasping angles were extracted from the rotation matrix of $\mathbf{T}_g'$ as follows:
\begin{equation}
\psi = \tan^{-1}\left(\frac{r_{32}}{r_{33}}\right), \quad 
\theta = \tan^{-1}\left(\frac{-r_{31}}{\sqrt{r_{32}^2 + r_{33}^2}}\right), \quad 
\phi = \tan^{-1}\left(\frac{r_{21}}{r_{11}}\right)
\end{equation}
where $r_{ij}$ represents corresponding rotation matrix elements. This approach enabled dynamic and precise grasping of planar objects in variable environments. To further enhance robustness during manipulation, we refined the detection algorithm to continuously update the estimated pose of the target object in real time. This iterative pose refinement persisted throughout the grasping process, leveraging ongoing visual feedback until the object was no longer visible to the camera. The loss of visibility typically occurred due to occlusions or positional constraints imposed by the camera's mounting on the end-effector. By maintaining pose updates until visual tracking was no longer feasible, the system demonstrated improved adaptability to environmental changes and object displacement, thereby increasing the reliability of grasp execution under practical constraints.

\section{Results and Discussion}\label{sec2}
Simulation studies and experimental validations were carried out to evaluate the effectiveness of the vision algorithm in detecting a textured cover. The inputs from vision algorithm were employed for manipulating motions of the mobile manipulator.
\subsection{Simulation study}
\begin{figure}[hbt!]
    \centering
\includegraphics[width=0.7\textwidth, height = 1.8 in]{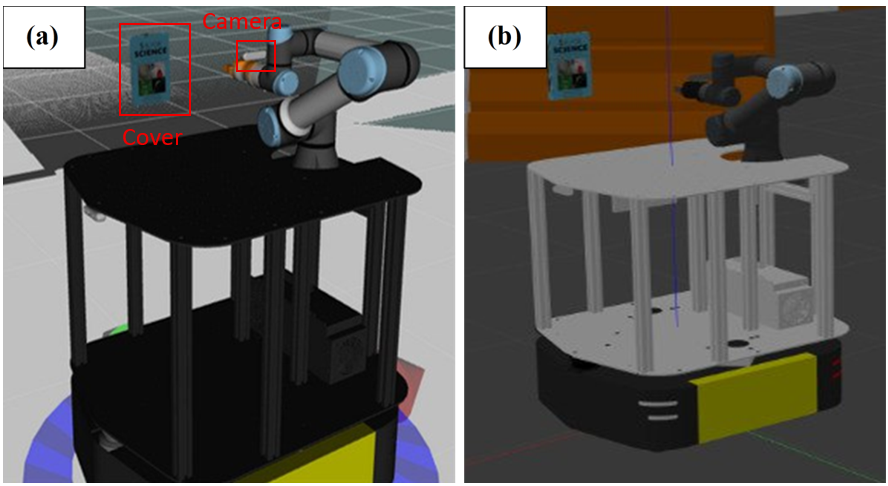} 
    \caption{Simulation environments of the mobile manipulator. (a) RViz visualization showing object detection and camera perspective. (b) Gazebo environment illustrating the robot setup for grasping experiments.}
   \label{fig.2}
\end{figure}

The implementation of the proposed object recognition and pose estimation algorithm was carried out in Robotic Operating System (ROS) environment integrating with OpenCV on an Ubuntu 20.04 platform, utilizing a 3.0 GHz Intel Core i7-7400 CPU with 16GB of system memory. 
We developed a simulation environment utilizing RViz and Gazebo, incorporating a mobile manipulator along with a textured cover of a book as shown in fig. \ref{fig.2}. The environments showcasing RViz and Gazebo environments are shown in figs. \ref{fig.2} (a) - (b).

\begin{figure}[hbt!]
    \centering
    \includegraphics[width=0.5\textwidth]{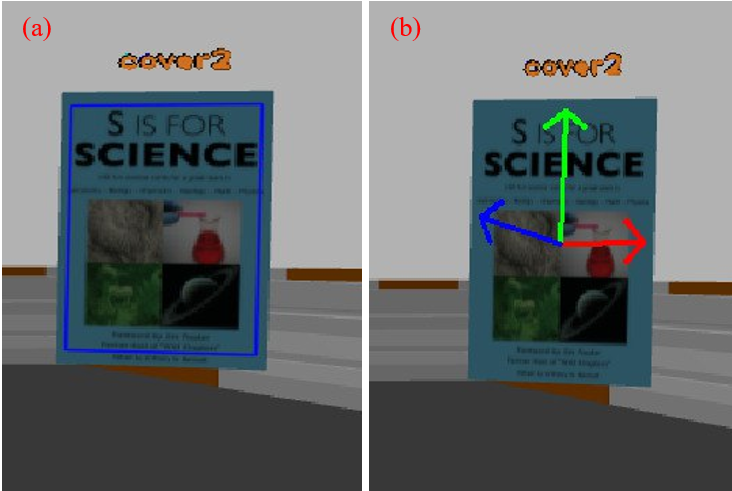}
    \caption{Vision algorithm outputs (a) Bounding box (b) Pose of object}
    \label{exp1}
\end{figure}

\begin{figure}[hbt!]
    \centering
\includegraphics[width=0.9\textwidth, height = 2.7 in]{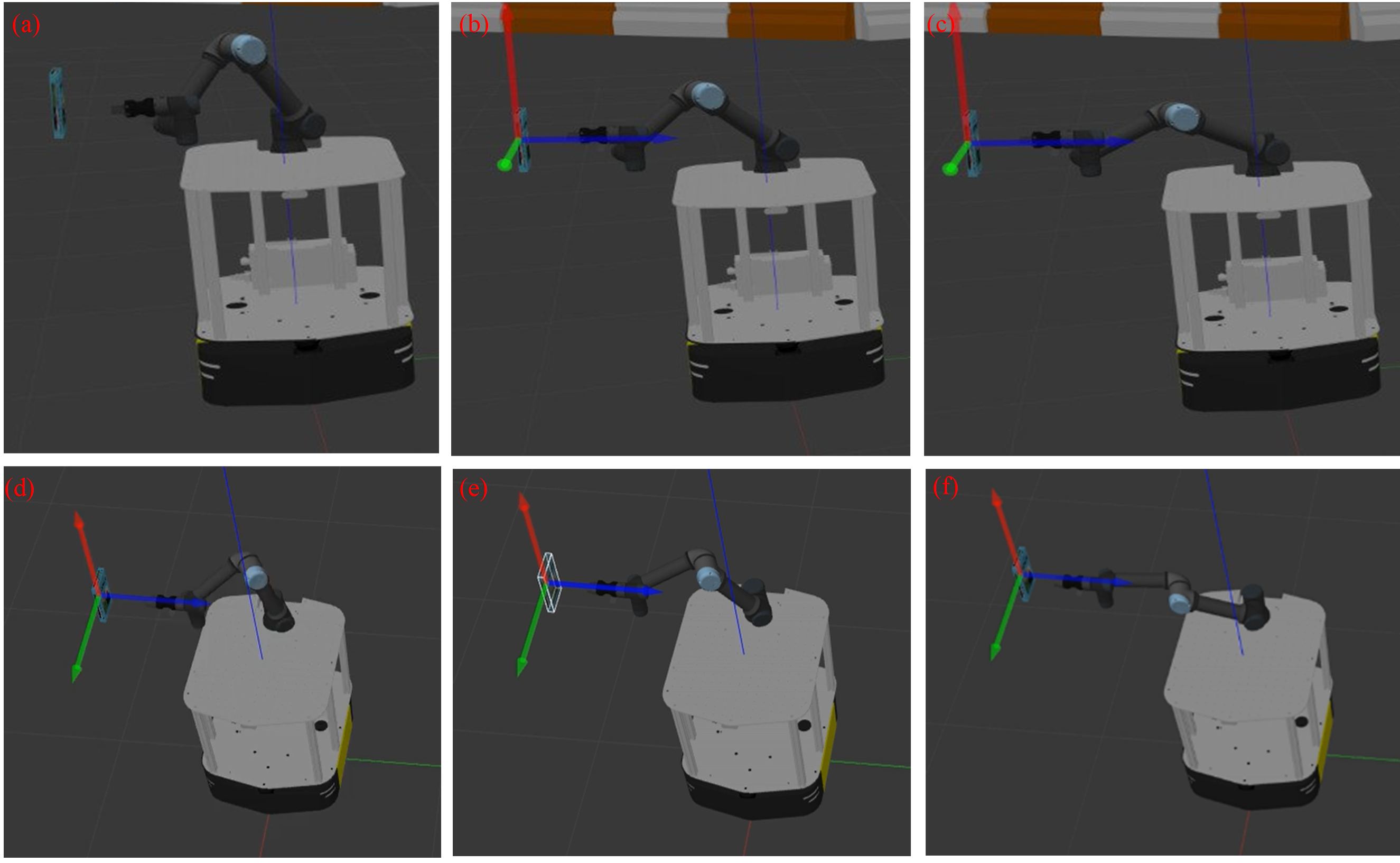} 
    \caption{Motions of manipulator to reach a book (a) - (c) Front view (d) - (f) Top view }
   \label{fig.3}
\end{figure}

%The mobile manipulator consisted of a non-holonomic mobile base with 4 omni-directional wheels, a UR5e arm, and a robotiq hande gripper. 
A RealSense camera attached to the end effector of the manipulator was used to detect continuously the pose of the textured front cover of a book. Once the pose was determined, the RRT* motion planning algorithm utilized this information to track and follow the book's movement with a grasp posture. Although trajectory planning was performed based on the predetermined grasp orientation of the end effector using ROS packages, these orientations were carefully finalized through workspace analysis using manipulator’s kinematic model to ensure avoidance of singular configurations. Beyond tracking, we also conducted complementary studies focused on active grasping of the book. In these experimentation scenarios, the system not only followed the book’s motion but successfully executed a grasp maneuver once the object reached a stable pose. These grasping trials (20 times) demonstrated the framework’s capability to transition from pose estimation and motion planning to physical interaction, validating its robustness in both tracking and manipulation tasks. This process was executed within the ROS MoveIt framework ensuring smooth and adaptive path optimization for precise object tracking and manipulation. The outputs of the vision algorithm are illustrated in Figs.\ref{exp1}(a) and (b), where Fig.\ref{exp1}(a) presents the detected bounding box around the object, and Fig.\ref{exp1}(b) depicts the estimated pose of the object. These visual results demonstrated the algorithm’s capability to accurately localize and orient the target within the scene, providing essential input for subsequent manipulation and planning tasks.
The screenshots of the motions of the manipulator in response to variations in the book's poses are shown in figs. \ref{fig.3}(a) - (f). These frames captured the adaptive motion planning and execution as the robotic arm adjusts its configuration to reach the target object. The manipulator, mounted on a stationary mobile platform, dynamically repositions its joints and end effector in accordance with the updated pose estimates provided by the vision system. The colored coordinate axes visible in each frame represented the estimated pose of the book, highlighting the system’s ability to track and align with the object throughout the approach. This visual evidence underscored the effectiveness of the integrated perception and planning framework in achieving precise and responsive object interaction.

\begin{figure}[hbt!]
    \centering
\includegraphics[width=1\textwidth, height = 1.5 in]{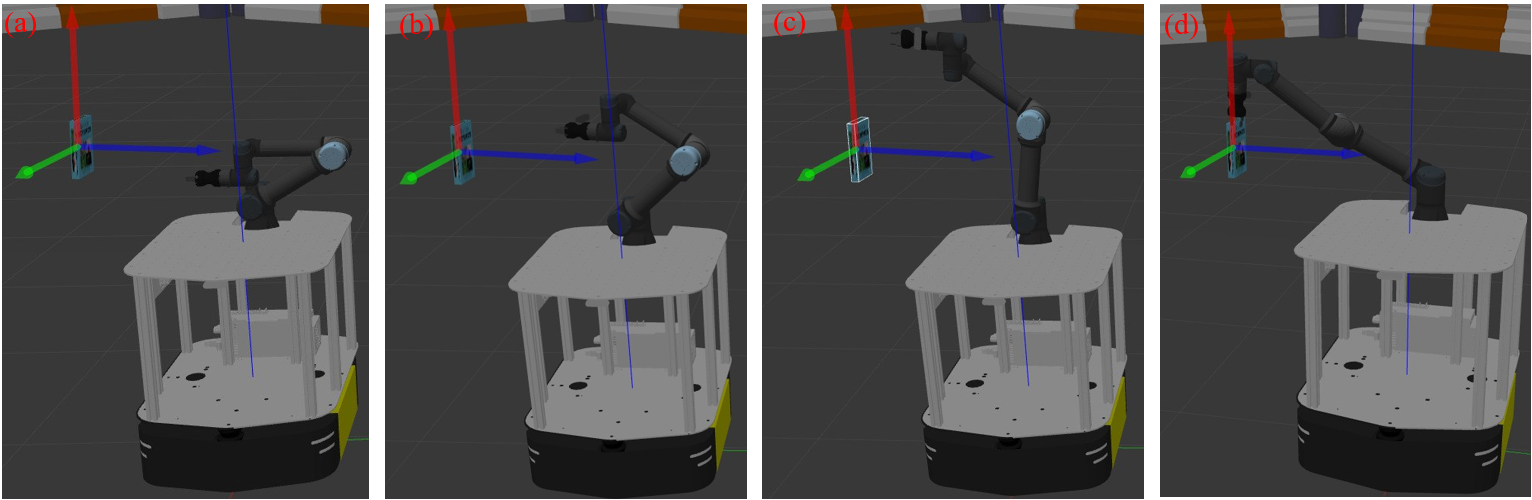} 
    \caption{Sequential motions of the mobile manipulator during the grasping task. (a) - (d) The arm approaches, aligns with, and securely grasps the textured book cover, guided by pose estimation from the vision system.}
   \label{fig.31}
\end{figure}

In addition to tracking the book’s pose, we conducted experiments focused on executing a grasp once the object was localized. Figs.\ref{fig.31}(a)–(d) illustrates the sequential motions of the manipulator as it performs a grasping task. Upon receiving the pose estimate from the vision system, the manipulator planned and executed a trajectory to approach and securely grasp the book. Unlike the tracking scenario, this sequence demonstrated a complete manipulation cycle from perception to physical interaction, where the robot transitions from pose estimation to a stable grasp.

\subsection{Experimental validation}
\begin{figure*}[hbt!]
    \centering    \includegraphics[width=0.4\textwidth]{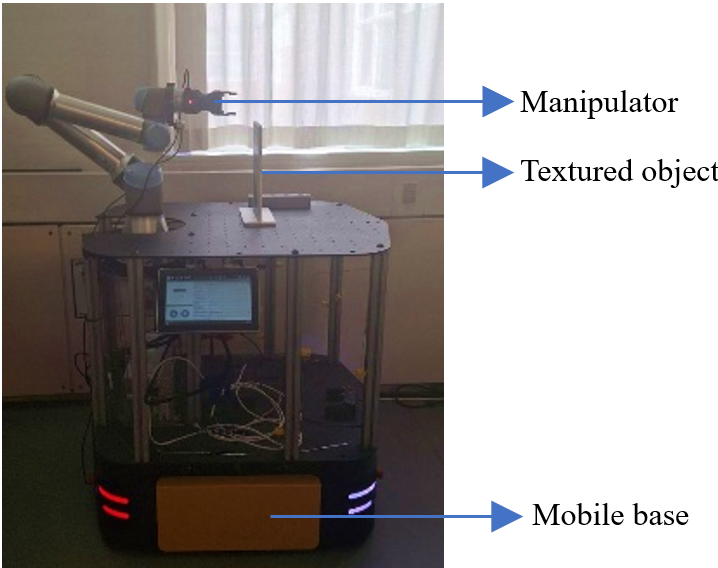}    \includegraphics[width=0.3\textwidth]{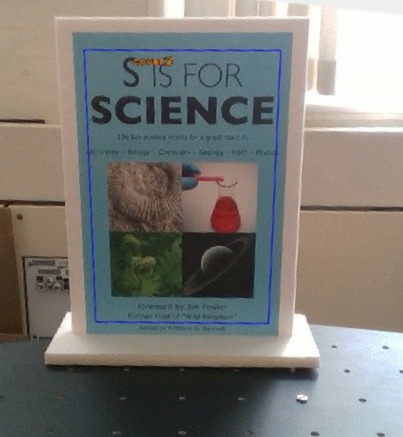}
    \caption{Experimental set up along with book cover}
    \label{ep1}
\end{figure*}

Fig. \ref{ep1} presents the experimental setup used to validate the robotic system’s object interaction capabilities. The configuration features a robotic manipulator mounted on a mobile base, positioned in front of a textured book cover. This setup replicated the conditions modeled in the simulation, allowing for a direct comparison between predicted and observed manipulator behavior. The book cover served as a visually rich target for pose estimation and tracking, enabling the system to test its perception and motion planning modules in a controlled environment. The manipulator’s initial posture confirmed its readiness to engage in object-following and grasping tasks.

\begin{figure}[hbt!]
    \centering
    \includegraphics[width=1\textwidth]{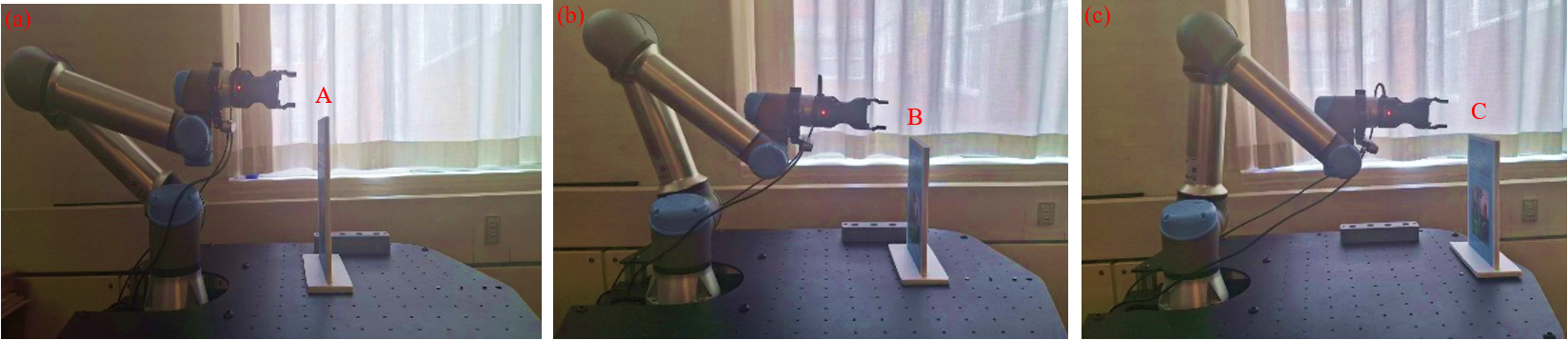}
    \caption{Manipulator following the book cover}
    \label{ep2}
\end{figure}

\begin{figure}[hbt!]
    \centering
    \includegraphics[width=1\textwidth]{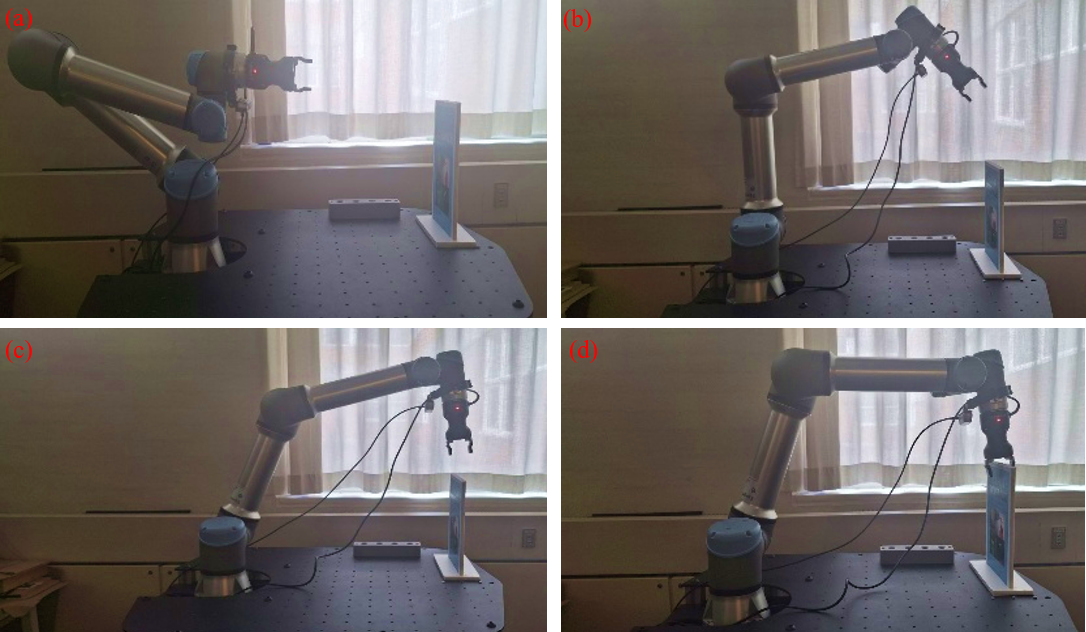}
    \caption{Manipulator grasping the book}
    \label{ep3}
\end{figure}

Fig. \ref{ep2} illustrates the manipulator’s ability to follow the textured object across a sequence of spatial configurations. The four subfigures (a–c) depict the robot arm dynamically adjusting its position as the book cover moves along a tabletop surface. This sequence demonstrates the integration of real-time visual feedback and motion control, allowing the manipulator to maintain alignment with the object. The tracking behavior is consistent with the simulation results, confirming the robustness of the pose estimation algorithm and the manipulator’s kinematic responsiveness. These observations validated the system’s capacity for continuous object monitoring in dynamic scenes. Fig. \ref{ep3} showcases the manipulator’s execution of a grasping maneuver, culminating in a successful grip of the book cover. Across subfigures (a–d), the robot transitions from a tracking posture to a grasp-ready configuration, ultimately securing the object. This sequence highlighted the effectiveness of the grasp planning module, which leveraged pose data and trajectory optimization to achieve stable contact. The grasping action confirmed that the manipulator can not only follow but also interact physically with the object, a critical capability for autonomous manipulation tasks. The consistency between simulated grasp strategies and experimental outcomes reinforced the reliability of the system’s control architecture.
Collectively, Figs. \ref{ep1}– \ref{ep3} demonstrated the successful translation of simulated robotic behaviors into real-world execution. The manipulator’s ability to detect, follow, and grasp a textured object validated the integrated perception and planning framework developed during simulation. 
%Add tracking accuracy, grasping success rate, errors, Like how fast it runs, how precise it is and how good it is at detecting the object.
\begin{table}[htbp]
\centering
\caption{Performance Metrics for Object Tracking and Grasping}
\begin{tabular}{|l|c|p{8cm}|}
\hline
\textbf{Metric} & \textbf{Value} & \textbf{Description} \\
\hline
Tracking Accuracy & 98.4\% & Percentage of frames with correct pose estimation ( $\pm$0.60 cm) before occlusion due to end-effector constraints. \\
\hline
Grasping Success Rate & 96.7\% & Ratio of successful grasps to total grasp attempts across varied object configurations and environments. \\
\hline
Pose Estimation Error & $\pm$0.53 cm & Average spatial deviation between estimated and ground-truth object pose. \\
\hline
Detection Latency & 71 ms & Average time per frame for object detection and pose update, enabling real-time responsiveness. \\
\hline
Detection Precision & 98.1\% & Proportion of correctly identified objects among all detections. \\
\hline
Detection Recall & 97.5\% & Proportion of actual objects successfully detected. \\
\hline
Runtime Performance & $\sim$68 FPS & Average processing speed during real-time tracking and grasping. \\
\hline
\end{tabular}
\label{tab:performance_metrics}
\end{table}

In order to record quantitative data, we conducted extensive trials using  the 
textured book cover positioned at various locations within the camera's field of view under diverse environmental conditions. Ground truth poses were obtained via fiducial markers and all estimated poses are expressed in the same coordinate frame for direct comparison. The tracking accuracy reached 98.4\%, indicating reliable pose updates until visual occlusion occurred due to end-effector constraints. Grasping success rate was recorded at 96.7\%, demonstrating robust manipulation capabilities even under partial occlusions and pose drift.
Pose estimation error remained within ±0.53 cm, reflecting high spatial precision suitable for fine-grained grasping tasks. The detection pipeline operated at an average latency of 71 ms per frame, enabling real-time responsiveness at 68 FPS. Detection precision and recall were 98.1\% and 97.5\%, respectively, confirming the system's effectiveness in identifying and localizing target objects under variable lighting and background clutter.
These results validate the system’s suitability for dynamic manipulation tasks in semi-structured environments, with strong generalization across object types and camera viewpoints. 
These experimental results underscore the system’s potential for deployment in practical applications such as automated sorting, assistive robotics, and mobile manipulation.

%The alignment between simulation and experimentation confirmed the fidelity of the modeling environment and the adaptability of the robotic platform to real-world variability.

%\section{Discussion}\label{sec12}

\section{Conclusion}\label{sec13}
This work commenced with the kinematic modeling of the manipulator using the Denavit–Hartenberg (DH) convention, followed by the determination of its inverse kinematics to ensure precise and adaptive manipulation capabilities. This work presented a comprehensive framework for enabling mobile manipulators to perform autonomous object tracking and grasping within self-driving laboratory environments. By integrating planar pose estimation with RGB and depth data, the system demonstrated robust perception capabilities, leveraging feature matching and homography estimation to accurately localize textured planar objects. The vision algorithm combined with adaptive grasp planning, enabled the manipulator to dynamically follow and interact with objects in both simulated and real-world settings.
This work implemented a framework for enabling mobile manipulators to track objects through planar pose estimation, leveraging RGB image and depth data.  Feature matching was conducted using FLANN’s K-d Tree Nearest Neighbor implementation and Bruteforce Hamming, while homography estimation was performed using RANSAC to ensure robust transformation. The resulting homography matrix facilitated the approximation of three orthonormal directional vectors on the object's surface via perspective transformation, which were subsequently used to estimate the planar pose.

Experimental validation confirmed that the manipulator could reliably track moving objects and execute stable grasping maneuvers once the pose was determined. These results underscore the effectiveness of the integrated perception, kinematic modeling, and motion planning pipeline in achieving precise and responsive manipulation. The consistency between simulation and physical trials further highlights the fidelity of the modeling environment and the adaptability of the robotic platform to dynamic conditions. Beyond object tracking, the system successfully transitioned to physical interaction, demonstrating its capability to grasp objects autonomously. Future work will focus on expanding the system’s grasping capabilities to include non-planar, transparent and deformable objects, enhancing real-time decision-making under uncertainty, and integrating multimodal sensing for improved environmental awareness. These advancements aim to further optimize autonomous experimentation and reinforce the role of intelligent robotics in next-generation chemical laboratories. While existing approaches address several challenges, achieving a universal solution that effectively handles dynamic environmental variations remains an open research problem. Our future work will focus on grasping the object and  expanding real-time experimentation in the self-driving laboratory, further optimizing the integration of robotics with adaptive grasping mechanisms for enhanced autonomy and efficiency.

%\bmhead{Acknowledgements}

%This work was supported by the Pioneer Center for Accelerating P2X Materials Discovery, CAPeX, DNRF grant number P3.

\section*{Declarations}

%Some journals require declarations to be submitted in a standardised format. Please check the Instructions for Authors of the journal to which you are submitting to see if you need to complete this section. If yes, your manuscript must contain the following sections under the heading `Declarations':

\begin{itemize}
\item Funding

This work was supported by the Pioneer Center for Accelerating P2X Materials Discovery, CAPeX, DNRF grant number P3.
\item Conflict of interest/Competing interests

No conflict of interest/Competing interests between authors.

\item Ethics approval and consent to participate

Not applicable
\item Consent for publication

All authors have provided their consent for publication of this manuscript.

\item Data availability 

Data supporting the findings of this study can be accessed through direct communication with the corresponding author.\item Materials availability

Not applicable
\item Code availability 

The code used in this study is available from the corresponding author upon reasonable request.

\item Author contribution

\end{itemize}

%\noindent
%If any of the sections are not relevant to your manuscript, please include the heading and write `Not applicable' for that section. 


\begin{thebibliography}{50}


\bibitem{ref1}
Zhou, J., Luo, M., Chen, L., Zhu, Q., Jiang, S., Zhang, F., ... Jiang,
J. (2025). A multi-robot–multi-task scheduling system for autonomous
chemistry laboratories. Digital Discovery, 4(3), 636-652.

\bibitem{n1}
Alexandra, Dobrokvashina, Sulaiman Shifa, Gamberov Timur, Hsia Kuo-Hsien, and Magid Evgeni. "New Features Implementation for Servosila Engineer Model in Gazebo Simulator for ROS Noetic"  In Proceedings of the International Conference on Artificial Life Science and Technology, vol. 28, pp. 153-156. ALife Robotics Co., Ltd., 2023.

\bibitem{n2}
Sulaiman, Shifa, A. P. Sudheer, and Evgeni Magid. "Torque control of a wheeled humanoid robot with dual redundant arms." Proceedings of the Institution of Mechanical Engineers, Part I: Journal of Systems and Control Engineering 238, no. 2 (2024): 252-271.

\bibitem{r1}
Paul, S.K., Chowdhury, M.T., Nicolescu, M., Nicolescu, M., Feil-Seifer, D. (2021). Object Detection and Pose Estimation from RGB and Depth Data for Real-Time, Adaptive Robotic Grasping. In: Arabnia, H.R., Deligiannidis, L., Shouno, H., Tinetti, F.G., Tran, QN. (eds) Advances in Computer Vision and Computational Biology. Transactions on Computational Science and Computational Intelligence. Springer, Cham. https://doi.org/10.1007/978-3-030-71051-4-10

\bibitem{ref9}
Viola, P., \& Jones, M. (2001). Rapid Object Detection using a Boosted Cascade of Simple Features. CVPR.


\bibitem{ref10}
Dalal, N., \& Triggs, B. (2005). Histograms of Oriented Gradients for Human Detection. CVPR.

\bibitem{ref11}
Girshick, R., Donahue, J., Darrell, T., \& Malik, J. (2014). Rich Feature Hierarchies for Accurate Object Detection and Semantic Segmentation. CVPR.

\bibitem{ref12}
Girshick, R. (2015). Fast R-CNN. ICCV.

\bibitem{ref13}
Ren, S., He, K., Girshick, R., \& Sun, J. (2015). Faster R-CNN: Towards Real-Time Object Detection with Region Proposal Networks. NeurIPS.

\bibitem{ref14}
Redmon, J., Divvala, S., Girshick, R., \& Farhadi, A. (2016). You Only Look Once: Unified, Real-Time Object Detection. CVPR.

\bibitem{ref15}
Liu, W., Anguelov, D., Erhan, D., et al. (2016). SSD: Single Shot MultiBox Detector. ECCV.

\bibitem{ref16}
Lin, T.-Y., Goyal, P., Girshick, R., He, K., \& Dollár, P. (2017). Focal Loss for Dense Object Detection. ICCV.

\bibitem{ref17}
Carion, N., Massa, F., Synnaeve, G., et al. (2020). End-to-End Object Detection with Transformers. ECCV.

%\bibitem{ref18}
%Zhu, X., Su, W., Lu, L., et al. (2021). Deformable DETR: Deformable Transformers for End-to-End Object Detection. ICLR.

%\bibitem{ref19}
%Zhang, H., Li, F., Liu, S., et al. (2022). DINO: DETR with Improved DeNoising Anchor Boxes for End-to-End Object Detection. arXiv.

\bibitem{ref18}
Jamali, M., Davidsson, P., Khoshkangini, R., et al. (2025). Context in Object Detection: A Systematic Literature Review. Artificial Intelligence Review.

\bibitem{ref19}
Wang, A., Chen, H., Liu, L., et al. (2024). YOLOv10: Real-Time End-to-End Object Detection. NeurIPS.


\bibitem{refSIFT1}
Piccinini, P., Prati, A., \& Cucchiara, R. (2012). Real-time object detection and localization with SIFT-based clustering. Image and Vision Computing, 30(8), 573-587.

\bibitem{refSIFT}
Vaithiyanathan, D., \& Manigandan, M. (2023, April). Real-time-based Object Recognition using SIFT algorithm. In 2023 Second International Conference on Electrical, Electronics, Information and Communication Technologies (ICEEICT) (pp. 1-5). IEEE.



\bibitem{pose3}
Brachmann, E., Krull, A., Michel, F., Gumhold, S., Shotton, J., \& Rother, C. (2014, September). Learning 6d object pose estimation using 3d object coordinates. In European conference on computer vision (pp. 536-551). Cham: Springer International Publishing.


\bibitem{pose4}
Tremblay, J., To, T., Sundaralingam, B., Xiang, Y., Fox, D., \& Birchfield, S. (2018). Deep object pose estimation for semantic robotic grasping of household objects. arXiv preprint arXiv:1809.10790.

\bibitem{pose1}
Wang, C., Xu, D., Zhu, Y., Martín-Martín, R., Lu, C., Fei-Fei, L., \& Savarese, S. (2019). Densefusion: 6d object pose estimation by iterative dense fusion. In Proceedings of the IEEE/CVF conference on computer vision and pattern recognition (pp. 3343-3352).

\bibitem{pose2}
Xiang, Y., Schmidt, T., Narayanan, V., \& Fox, D. (2017). Posecnn: A convolutional neural network for 6d object pose estimation in cluttered scenes. arXiv preprint arXiv:1711.00199.

\bibitem{c1}
Sultanov, Ramir, Shifa Sulaiman, Tatyana Tsoy, and Elvira Chebotareva. "Virtual collaborative cells modeling for UR3 and UR5 robots in Gazebo simulator." In Proceedings of the 2023 International Conference on Artificial Life and Robotics, pp. 149-152. 2023.

\bibitem{c2}
Sultanov, Ramir, Shifa Sulaiman, Hongbing Li, Roman Meshcheryakov, and Evgeni Magid. "A review on collaborative robots in industrial and service sectors." In 2022 International Siberian Conference on Control and Communications (SIBCON), pp. 1-7. IEEE, 2022.


\bibitem{ref2}
Angelopoulos, Anastasios, Matthew Verber, Collin McKinney, James Cahoon, and Ron Alterovitz. “High-Accuracy Injection Using a Mobile Manipulation Robot for Chemistry Lab Automation.” IEEE/RSJ International Conference on Intelligent Robots and Systems (IROS), 2023. 

\bibitem{ref3}
Fizet, Julien. “Lab Automation with Mobile Robots.” Antdriven Insights, 2025. https://insights.antdriven.com/lab-automation-mobile-robots

\bibitem{ref55}
ANTdriven. (n.d.). Revolutionizing lab automation with mobile robots. ANTdriven Insights. https://insights.antdriven.com/lab-automation-mobile-robots

\bibitem{ref4}
Robotnik Automation. “Uses of a Mobile Manipulator Robot: Beyond Robotic Arms.” Robotnik Blog, 2025. https://robotnik.eu/uses-of-a-mobile-manipulator-robot-beyond-robotic-arms/

\bibitem{ref5}
CRL Solutions. “Telemanipulators for Radiopharmaceutical Laboratories.” CRL Technical Overview, 2024. https://crlsolutions.com/products/telemanipulators/

\bibitem{ref6}
PS Lift. “Vacuum Manipulators in the Chemical and Pharmaceutical Industry.” PS Lift Product Guide, 2024. https://en.ps-lift.com/product/vacuum-manipulator-in-chemical-and-pharmaceutical-industry/

\bibitem{ref7}
Dalmec North America. “How Industrial Manipulator Solutions Benefit the Chemical Industry.” Dalmec NA Blog, 2024. https://www.dalmec-na.com/blog/how-industrial-manipulator-solutions-benefit-the-chemical-industry/

\bibitem{ref8}
Debijadi, M. L. (2023). Simulation design of a robotic mobile manipulator
for Material Acceleration Platforms (Master’s thesis, Aalborg University).
Aalborg University Student Projects.




\bibitem{3}
M. Muja, D.G. Lowe, Fast approximate nearest neighbors with automatic algorithm configuration,
in International Conference on Computer Vision Theory and Application VISSAPP’09)
(INSTICC Press, Lisboa, 2009), pp. 331–340

\bibitem{4}
M.A. Fischler, R.C. Bolles, Random sample consensus: a paradigm for model fitting with
applications to image analysis and automated cartography, Commun. ACM 24(6), 381–395
(1981)



\end{thebibliography}
\end{document}